# Design and Implementation of English To Yorùbá Verb Phrase Machine Translation System

**Benjamin Ajibade Ayoade and Safiriyu Eludiora**
Department of Computer Science and Engineering,
Obafemi Awolowo University, Ile Ife, Nigeria.

## Abstract

The Yorùbá group has diverse language speakers across the world, translating the language to other widely spoken languages must be emphasized. We aim to develop an English to Yorùbá machine translation system which can translate English verb phrase text to its Yorùbá equivalent. Words from both languages Source Language and Target Language were collected for the verb phrase group in the home domain. The lexical translation is done by assigning values of the matching word in the dictionary. The syntax of the two languages was realized using Context Free Grammar, we validated the rewrite rules with finite state automata. Human evaluation method was used and expert opinion scored. The evaluation shows the system performed better than that of sampled Google translation with over 70% of the response matching that of the system's output.

## 1 Introduction

The advancement in Natural language Processing (NLP) can be attributed to recent improvements in the strategy and techniques of large data collection, archiving, analysis, and visualization. NLP began in the '50s as machine translation (MT), intended to aid in code-breaking during World War II although the translations were not successful, these early stages of MT were necessary stepping stones on the way to more sophisticated technologies (Zhang, 2018; Quinn, 2017).

Machine Translation, therefore, is a more classical term, that is classified as a subfield of artificial intelligence that specifically uses computer software for automatic text translation in the absence of a human translator or where humans sparingly participate in the translation process. Machine Translation is highly essential in areas of human needs, while it is gradually taking over from the commonly used human translator its ever-improving translation techniques promise to be a faster and cheaper alternative. If communication is an integral part of human coexistence then language translation is the engine for its success.

This field of interest is driven by the rapid rise in the area of global information exchange, of which language has been a common barrier in global villages. Hence translation is an important factor where there exists a language barrier or extreme differences. According to Awoyale and Bamba (2015) in West Africa today, the continent's sub-region alone is home to native speakers of over 500 languages and one of the major African languages is Yorùbá (where there are 19,380,800 native speakers and between 45 and 55 million speakers across the world). Despite the population of speakers, Yorùbá is still considered as a low resource language (for which few language resources exist), making it very difficult for the development of more advanced models such as the Neural Machine model that requires large volumes of data. With the number of speakers, translating the language to other widely spoken languages was not initially emphasized. However, recent linguistic researchers are taking up the challenges by giving more attention (as compared to the high-resource language of the Western World).

The area of interest is machine translation (MT). MT is defined by Kishore *et al.* (2002) as the use of a computer to translate a message, typically text or speech, from one natural language to another. At its basic level, machine translation performs simple substitution of words in one natural language for words in another (Ruchikal and Gupta, 2014). MT systems also are often needed for translating literary works from any language into native languages, this breaks language barriers by making available rich sources of literature to people across the world (Ruchikal and Gupta, 2014; Tsegay and Azath, 2020).

Essentially, there are two major types of translation, which can be categorized as the partial translation and the full translation (Novianti, 2012), full translation is when every part of the source language text is replaced by the target language text material, while in partial translation some part of the source language text is left translated. In any type of translation that could be implemented, the translator's task is to represent the meaning of source text in the clearest and most acceptable form, the technique of translation is always secondary to the understanding of the source text (Aresta et al. 2018).

A phrase is a part of a clause or sentence (usually a single grammatical unit) with no complete sense. A verb phrase on the other hand usually consists of a verb and a preposition phrase or noun phrase. There has been significant work on English to Yorùbá (E-Y) phrase translation systems such as noun phrase translation (Abiola *et al*., 2014). This research work focuses on the E-Y verb phrase translation. The aim of this project is to develop English to Yorùbá machine translation system which can translate verb phrase text in English to its Yorùbá equivalent.

The objectives are to

1. collect words from both languages (SL and TL) for verb phrase groups.
2. design the translation process model of the two languages.
3. implement the model in (2).
4. evaluate the model implemented in (3).

Different models on language translation systems already exist but non-treated verb phrases, this research handles verb phrase as a contribution to knowledge.

# 2 Literature Review

## 2.1 Yorùbá: A Brief Introduction

The Yorùbá is today one of the three main ethnic groups that make up Nigeria. Yorùbá people are a large group or ethnic nation in Africa, and the majority of them speak Yorùbá, a language spoken natively by about thirty million people in Nigeria and in the neighboring countries of the Republic of Benin and Togo (ProjectSolutonz, 2019). Yorùbá has 3 syllables C, CV, and the syllabic nasal N. It has 17 consonant phonemes, /b,f,m,t,d,s,l,r,ʤ,ʃ,j,k,g,k͡p,g͡b,w,h/. (Jolaade, 2016). Yorùbá is also a tonal language with three tones, "high" "mid "and "low". According to Eludiora (2014), the high tone is indicated by an acute accent á, é, ẹ́, í, ó, ọ́ and ú while the mid-tone is not marked and the low tone is marked with a grave acute (à, è, ẹ̀, ì, ò, ọ̀ and ù). The mid-tone is generally unmarked except there might be ambiguity or confusion.

### i) Yorùbá sentence

The Yorùbá language clearly follows the SVO (Subject-Verb-Object structure) sentential word order. For example, "Ade pa ewúré" means "Ade killed a goat", "Ade" as the Subject, "killed" as Verb (transitive) and goat (Object). The basic structure is as shown

S → NP (Aux) VP NP

where S is the sentence, NP is a noun phrase and VP is a verb phrase, N is a noun and V is the verb.

NP → N,

VP → V

NB: The NP serves as the Object as well as Subject.

In a case when the verb is intransitive (does not require an object, it comes with the structure;

S → NP (Aux) VP

where S is the sentence, NP is a noun phrase and VP is the verb phrase, N is noun and V is verb.

S → NP (Aux) VP

NP → N

VP →V

### ii) Yorùbá Transitive Verb Phrase

A transitive verb phrase consists of a transitive verb and an object or a modifier. For example - "rí kìnú kan" which means "see a lion".

The verb "rí" comes with a Noun-phrase "kìnú" as a complement accompanied with an indefinite article "a". The phrase structure for such can be represented as

VP →V NP DET

Where V is the verb, DET is the article and NP is Noun-phrase.

#### iii) Yorùbá Intransitive Verb Phrase

This verb phrase includes an intransitive verb but does not include a Noun Phrase as an object to complement the verb. For this example, the phrase" fò kia " means "jump quickly", the Verb "fò" is the Verb-phrase (VP) and followed by an adverb "kia". In this case, the intransitive Verb can be followed by an adjunct/modifier.

VP →V AdvP

Where V is verb, AdvP is an adverbial phrase and Adv is an Adverb,

VP →V

AdvP →Adv

However, it is important to note that, Noun Phrase in Yorùbá is quite different from English as the determiners and adjectives, follow the noun, making Yorùbá noun phrase head initial. Yorùbá has poor morphological processes, such as inflection; instead, it uses syntax to convey the grammatical meaning. However, Unlike English Yorùbá can express its inflection in nouns and verbs with the help of auxiliaries and other words. A common practice is when a native speaker adds a word before the noun instead of inflecting it.

### A. Machine Translation

Machine Translation (MT) is a mode of translation in which computer applications are delegated the task of accepting source language (usually text and audio), process the source language according to the instructions and data provided, generating a target language equivalent of the source language and unaltered in meaning as required by the user.

Compared to human translation technique, MT also provides

1. a substitution of one language for another language,
2. study of how to form and structure words,
3. syntax which is the rules over the sequence of words combination i.e., how words are to be combined to form phrases and sentences,
4. semantics which is the branch of linguistics that deals with the meaning of words.

Human translator interprets and analyses all conditions within the text to understand how each word influences the context of the text. An expert (knowledge of the syntax and the semantics of the language of interest) is required in source and target language to achieve good result. The Auto translator explicitly simulates the expert knowledge of the structure and rules of the languages with a computational model. Auto-translation has challenges, for instance, automated translators might find it difficult to interpret some context based on the type of system.

Also, their exist challenges in the dataset used, model training. If implemented correctly it promises to improve economies of scale when translating in domains suited to MT.

### B. Rule-Based Machine Translation (RBMT)

The rule-based machine translation systems are built on dictionaries and linguistic rules and depend on the quality and volume of translated text (both the TL and the SL) in the database; say large corporal. This method is applicable when large collections of rules, developed manually over time by linguists that are both TL and SL experts. The rules are then hand-coded, while the system is delegated to find a matching translation and combines them to give a target text. RBMT is a good approach for MT engines of language with lots of grammar (Eludiora, 2014) which Yorùbá language is one.

It is also a good practice for low resource language. Users can improve translation quality by adding terminology into the translation process by creating a user-defined dictionary and improving on the rules generated.

### C. Statistical Machine Translation (SMT)

In the mid-20th century, Warren Weaver came up with the proposed theory of "cryptanalysis" (Quinn, 2017); given enough data, patterns generated should be applicable for other newer translations. The Statistical MT usually makes use of very large bilingual corpora for efficiency (Benyamin *et al*., 2018). Finding this type of volume can be specific to some notable languages (the application of SMT is rare to achieve in a low resourced language). The bilingual corporal gets trained until it learns the language's pattern, such pattern is expected to be efficiently reproducible. Brown *et al.* (1990) showed phrase-based SMT models to be often defined with the log-linear framework as shown in the equation below.

$$P(t|s) = \frac{exp\ (\sum_{m=1}^{M} \lambda_m\, h_m(t,s))}{\sum_i exp(\sum_{m=1}^{M} \lambda_m\, h_m(t,s))} \quad (1)$$

where $h_m(t,s)$ is a feature function and $\lambda_m$ is the weight.

### D. Neural Machine Translation

The Neural Machine Translator (NMT) is based on Deep Neural Network, an architecture designed to imitate the working principle of neurons of the human brain. LSTM (Long Short-Term Memory) and GRU (Gated Recurrent Unit) has proven to be efficient over common neural network algorithm such as ANN as they learn on long term dependencies quickly. This method can be used to map source sequence (input) to target sequence(output) via an encoder and decoder, LSTM and GRU being the basic building block. The encoder extracts a fixed-length representation from a variable-length input sentence, and the decoder generates a correct translation from this representation (Cho *et al.,* 2014). Such a complex system would be time-consuming to learn new language pairs.

As the computational demand of training NMT becomes more widely available and more research is performed, these gaps (accuracy and time optimization) will hopefully be filled in**.**

### E. Related Review

Abiola *et al* (2014) worked on the computational model of the English to Yorùbá (E-Y) noun-phrase translation system. Their project is resolving the restriction of computers to only those who understand the English language in Nigeria which consequently demoting the development of indigenous language. Rule-based translation model (CGG) was the approach used. The model was tested on one hundred and 60 randomly selected noun phrases from daily news and was used to test the translation. They look forward to using an improved machine language technique in their future work.

Eludiora (2014), implemented and designed English to the Yorùbá Machine Translation System. The authors are aiding the use of the Yorùbá language on computer systems. Transfer Rule-Based, the rewrite rule was verified using NL Toolkits (and implemented using python language).

## 3 Methodology

The translation process was modelled using a Unified Modelling Language (UML).

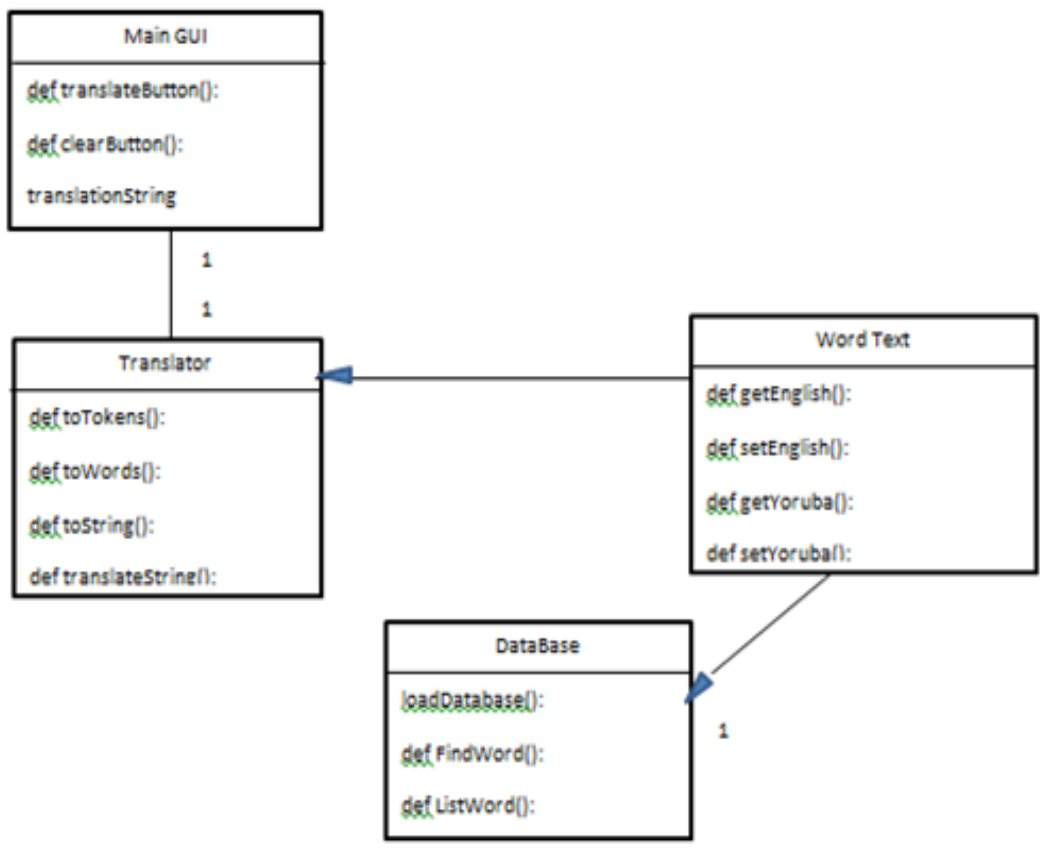

Figure 1: The Unified Modelling Language (UML) Diagram

A rule-based approach was used in the system's architectural design; using principles and rules developed based on the two languages. These rules guide the translation from the source language to the target language as shown in the class activity diagram in fig 3.1, the architecture of the system was designed based on the system's capability to tokenize input text (source language) i.e., English language, the token is then re-arranged according to the rewrite rules.

After which it has searched the dictionary for lexical translation of the token; and then gives the equivalent translation in the target language (Yorùbá language). The output of the system is displayed by the graphical user interface.

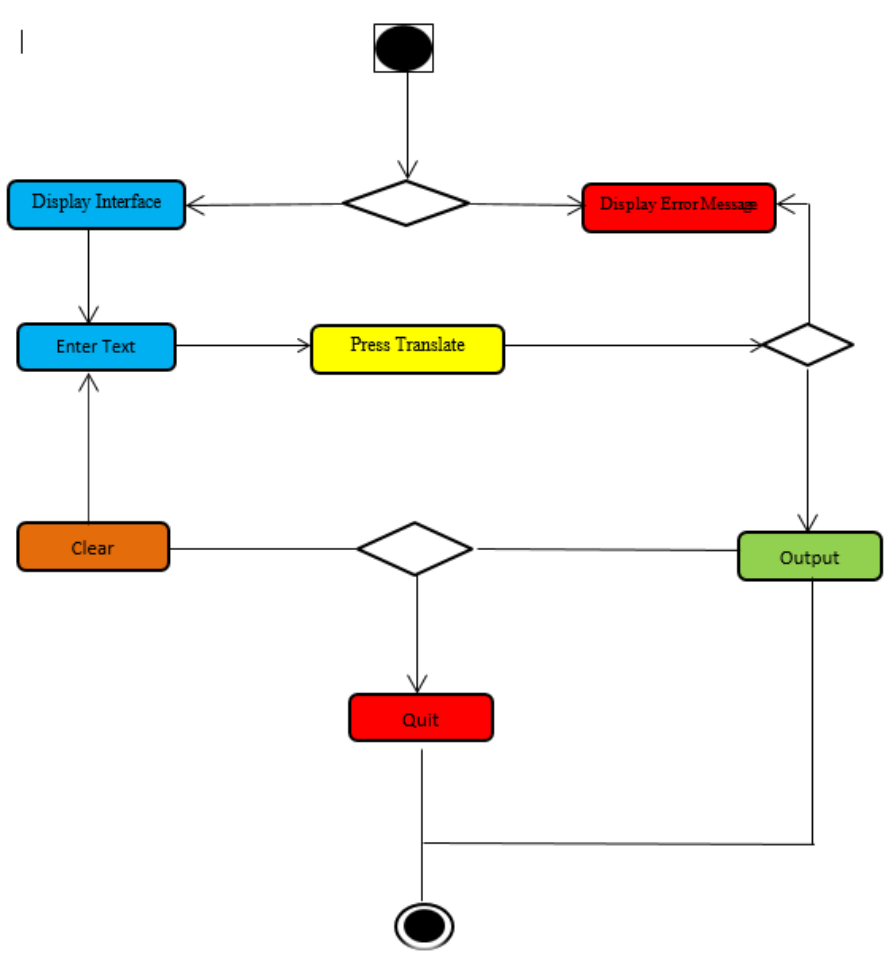

Figure 2: Showing the Class Activity

## 3.1 Data Collection and Storage

The data collected are five parts of speech (nouns, pronouns, adjectives, verbs, prepositions) and their respective Yorùbá equivalent stored in a python dictionary syntax. The data collected is limited to the home domain i.e., the words used and translated are familiar words that we often use at home. During data collection resources consulted include the internet, books, journals, and newspapers as well as the natives, since there exists no ready-made resources system library (Yorùbá is one of the many under-resourced African languages).

The collected words are group and represented using dictionary for retrieval. The sample of the data is shown

```
nouns = {'father': 'bàbá', 'mother':'iya','boy':'ọmọdọkúnrin',

det  =  {'the': 'náà', 'a': 'kan'}

verbs = {'gave':'fún','give':'fún','wash': 'fọ','washed':'fọ',
          'kill':'pa','fight':'já','are': 'wa', 'sat': 'jókò',
preps = {'in': 'nínu', 'to': 'sí', 'for': 'fún', 'with': 'pẹ̀lu

adjectives = {'cold': 'tútù', 'small': 'kékeré', 'big': 'nlá',
```

Figure 3: Sampled dictionary

## 3.2 Translator Engine

The development process was divided into stages, the Graphical User Interface (GUI) and the Translator Engine. The translator engine designed (database plus rules) using a monolithic architecture i.e. the components are developed as a single unit. The GUI is designed in three layers; the first layer accepts input from the user while the other two layers are used to output the result. The lexical translation is done by a search through the diction for a match after the sentence has been tokenized. The values of the corresponding text are returned, rearranged, and combined using these rules. The concatenated words are sent back to the GUI in order to display the grammatically correct Yorùbá equivalent.

## I. Grammar

The type of grammar used in this work is context free grammar. The verb phrase analysis depends upon knowing which theory one obtains in the context. Using the formal language theory, Let suppose that $\beta$ = (V, $\Sigma$, $\rho$, $\gamma$) is a 4-tuple context-free grammar, where V is a set of nonterminal symbols, $\Sigma$ is terminal symbol, $\rho$ is production rule and $\gamma$ is the start symbol which is an element of V. Let L($\beta$) be the function of language generated by $\beta$ over the finite alphabet $\Sigma$; A typical rule is A → $\omega$ where A is a symbol of V and $\omega$ is a string element in the language. David (2020) shows the language generated by $\beta$ is L($\beta$) over the alphabet $\Sigma$.

$$L(\beta) = \{\omega \in \Sigma \mid \gamma\} * \Rightarrow * \beta\ \omega \qquad (2)$$

This contains terminal symbols with start symbol $\gamma$ and applying the suitable production rules $\rho$.

We represent the base sentence as a typical simple sentence which consists of the non-terminal noun-phrase (NP) and verb phrase (VP) where S is a single nonterminal symbol. Where VP is a verb phrase, V is the verb, N is a noun, NP is a noun phrase, ADJ is an adjective, DET is determinant, P is a preposition, PP is a prepositional phrase.

***Start Rule***

**S** → NP VP
**NP** →N
VP → V

This grammar above takes any noun and any verb and shows how to combine them to form a sentence. In this work we focus on the verb phrase. As such we need to improve on our verb phrase rule a bit more. Prepositional phrases can be handled with the rule as well, by combining our free structure rules for noun phrases and verb phrases,

### 1) Source Language Production Rule

From the start rule structure derive a simple but more complete phrase structure grammar.
Now given the Verb phrase as a single nonterminal left-hand symbol VP, we can parse the 6 right-hand rules (terminal and nonterminal).

**VP** → V,
**VP**→ V NP,
**VP** → V ADJ N PP,
**VP**→ V DET ADJ N PP,
**VP**→ V DET ADJ N P NP,
**VP** → V DET ADJ N P DET N.

From the above rules, according to the scope of this work, all non-terminal symbols is would be:

***Verb Phrase:***
**VP** → V NP,
VP → V NP PP,
VP →V
***Noun Phrase:***
**NP** → N,
NP → DET N,
NP →DET ADJ N
***Prepositional Phrase:***
**PP** → P NP

Some verb phrases and structures are shown below,

V ⇒ | eat |
V NP ⇒ | eat | | the meat |
V PP ⇒ | eat | |on the table|
V NP PP ⇒ | eat | |the meat| |on the table|

While the terminal symbols are :
**V** ⇒ |ate|,
**N** ⇒ |meat|, |table|,
**ADJ** ⇒ |fresh|,
**DET** ⇒ |a| |the|,
P ⇒ |on|

### 2) Target Language Production Rule

From the start rule, given the verb phrase as a single non terminal left hand symbol VP. Generally, the phrase structure Rule for all the VPs in Yorùbá would be VP→ V (NP) (PP) (AdvP). For the for this work, we derived 4 right-hand production rules (terminal and non-terminal) for Yorùbá verb phrase as given below,

**VP** → V,
**VP** → V NP,
**VP** →V N ADJ,
**VP** → V N ADJ DET PP.

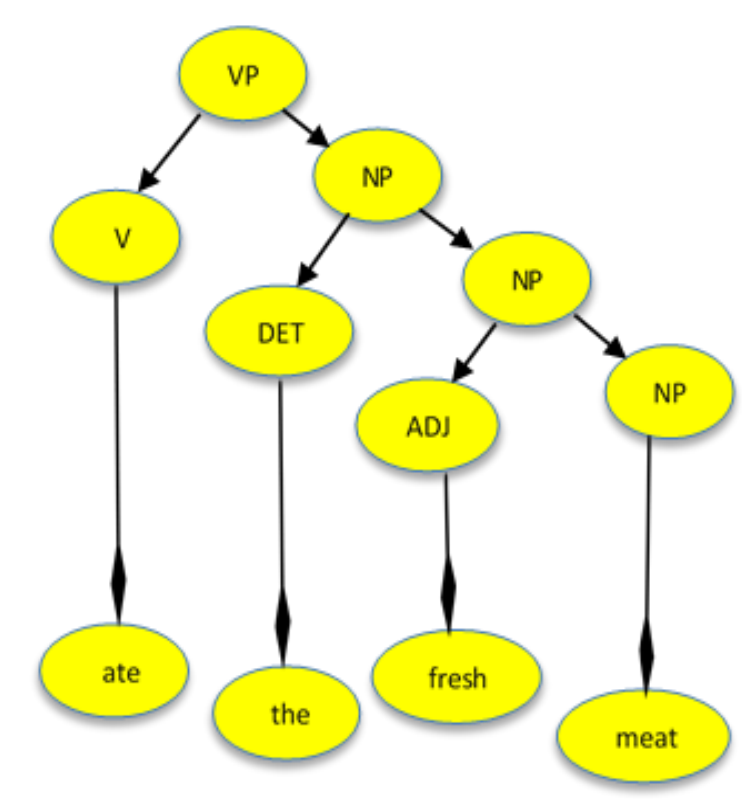

Figure 4: The VP Parse Tree for the Source

The terminal and non-terminal symbols are maintained but there are changes in the language, While the terminals of the Yorùbá production rule remains the same with the source language (English) production rules, there exists a slight variation in some of its structure.

V ⇒ | je |
V NP ⇒ | je | | eran naa |
V PP ⇒ |je| |lori tabili|
V NP PP ⇒ | je | |eran tutu naa| |lori tabili|
The terminals of the verb phrase representation in Yorùbá language are as listed,
**V** →| je |,
**N** → |eran|, |tabili|
**ADJ** → |tutu|,
**DET** → |naa|,
**PP** → |lori|

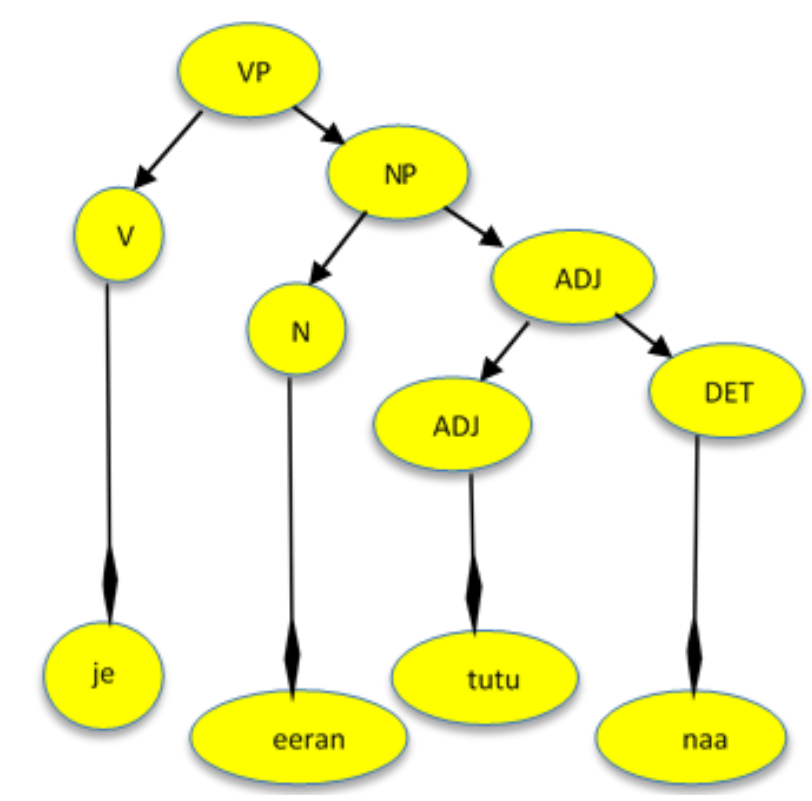

Figure 5: VP parsed tree for target language

Note that the rules are not limited to the Yorùbá verb phrase structure but within the scope of this work.

## II. Finite State Automata (FSA)

The finite automaton is a mathematical model used to compute how both systems changed states based on the input supplied, the FSA model was simulated using JFLAP in fig 6 and 7.

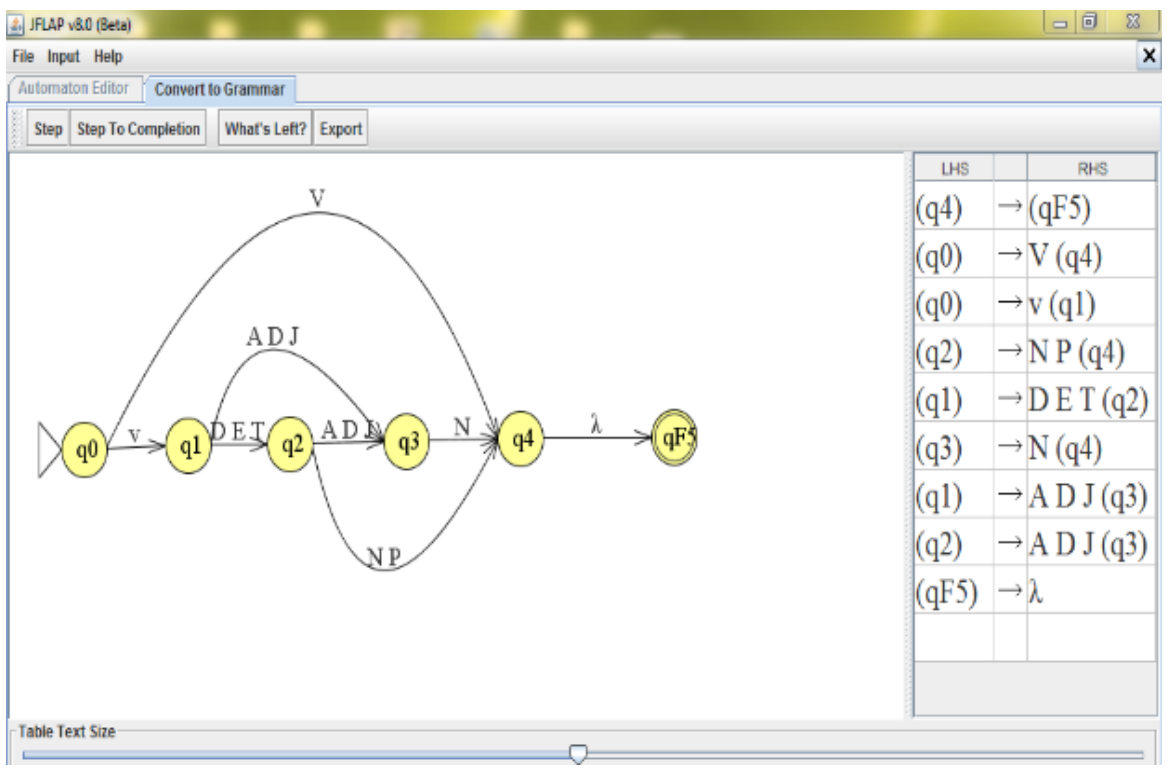

Figure 6: Showing the FSA model of Source Language

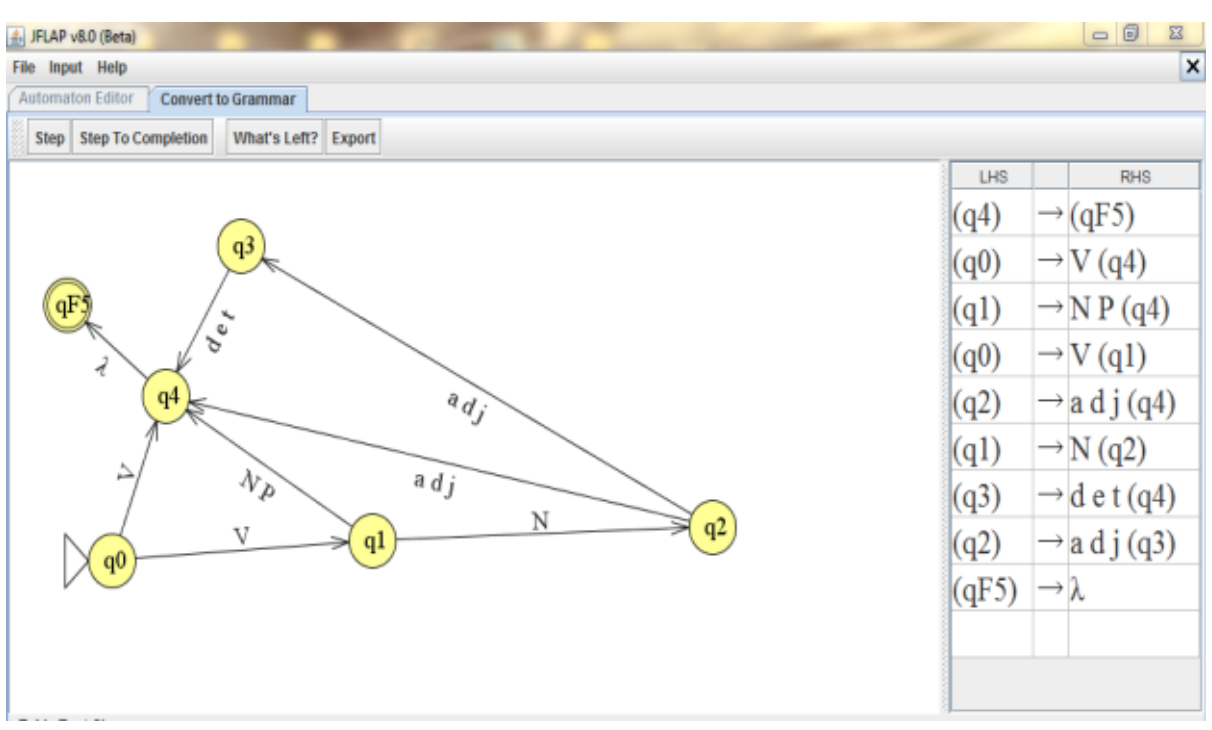

Figure 7: Showing the FSA model of Target Language

## 4 System Implementation

Python programming language was used for the software coding and the interface of the machine is designed using a module named Tkinter[1]. The translation process is based on the grammar built in the program code which follows the rewrite rules. The graphical user int GUI gives a user-friendly interaction. The output section is divided into two parts: the first part contains a text box that has the direct diction-based translation in the Yorùbá language and the second text box which contains rule-based translated words.

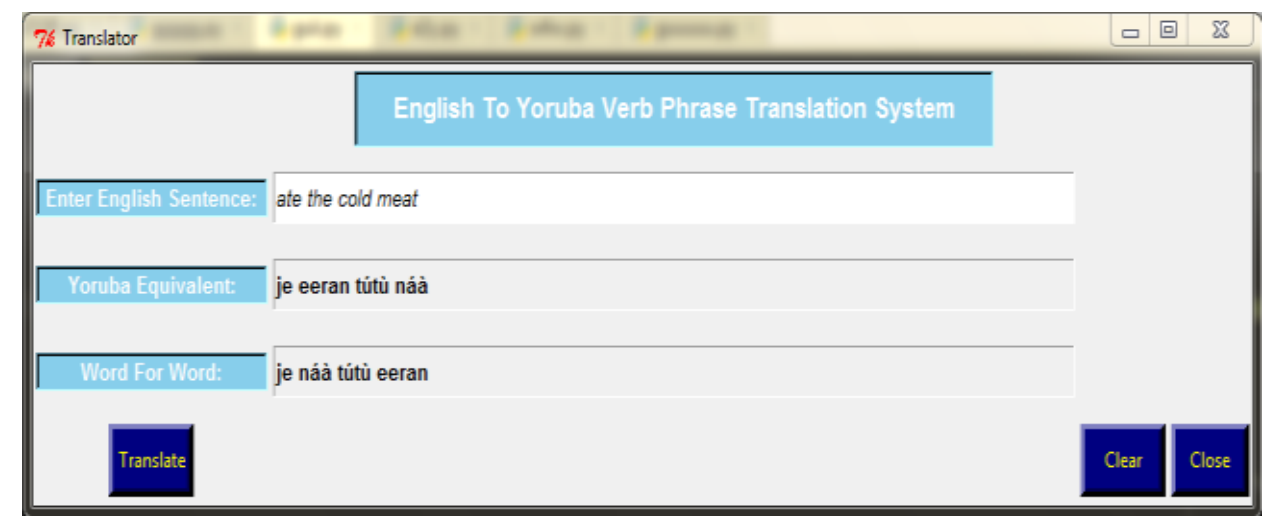

Figure 8: Translation Sample

***Create widget:*** the widget method links the user interface to the engine using the get () method.

***Initialize the empty list*****:** a list that holds text having the same part of speech (POS).

***Initialize empty string*****:** initialize an empty string to hold the converted/concatenated text and another to hold Word for Word (w4w) translate.

***Initialize the dictionary*****:** dictionary stack as database.

> ***def intoTokens***: returns a token from an input text.
>
> ***def group*****:** grouping the tokens to their respective part of speech by iterating through the tokenized text and appending to the list created.
>
> ***def convert***: iterating through the list if a word has a match in the initiated POS list then append the target language equivalent. Iterating through the initialized list for each word using len(list)+1 and starting from the last appended value.
>
> ***def w4wconvert*****:** word for word(w4w) method iterate through the list, if a word has a match in the initiated POS list then append the target language equivalent. Return the dictionary value of tokenized text, and then concatenate the strings.

**Output:** prints the output of the convert () and w4w convert method to the widget using the get().

Figure 9. Showing a quick the step-by-step methods used for the word for word and rule-based output

[1] https://github.com/BenAji/VerbPhrase

## I. Evaluation

The human evaluation method is used for the evaluation of this system. Although this method is time-consuming, it is very extensive. The system output and google translator output were compared with the opinion of the respondent (language expert) using some random examples from the data collected. In total we have 70 responses, the opinion shows that 69% of respondent translations tally with the E-Y VP translation, 29% goes with the google translation while we have 1% having a totally different opinion. The responses were collected using Microsoft excel online form.

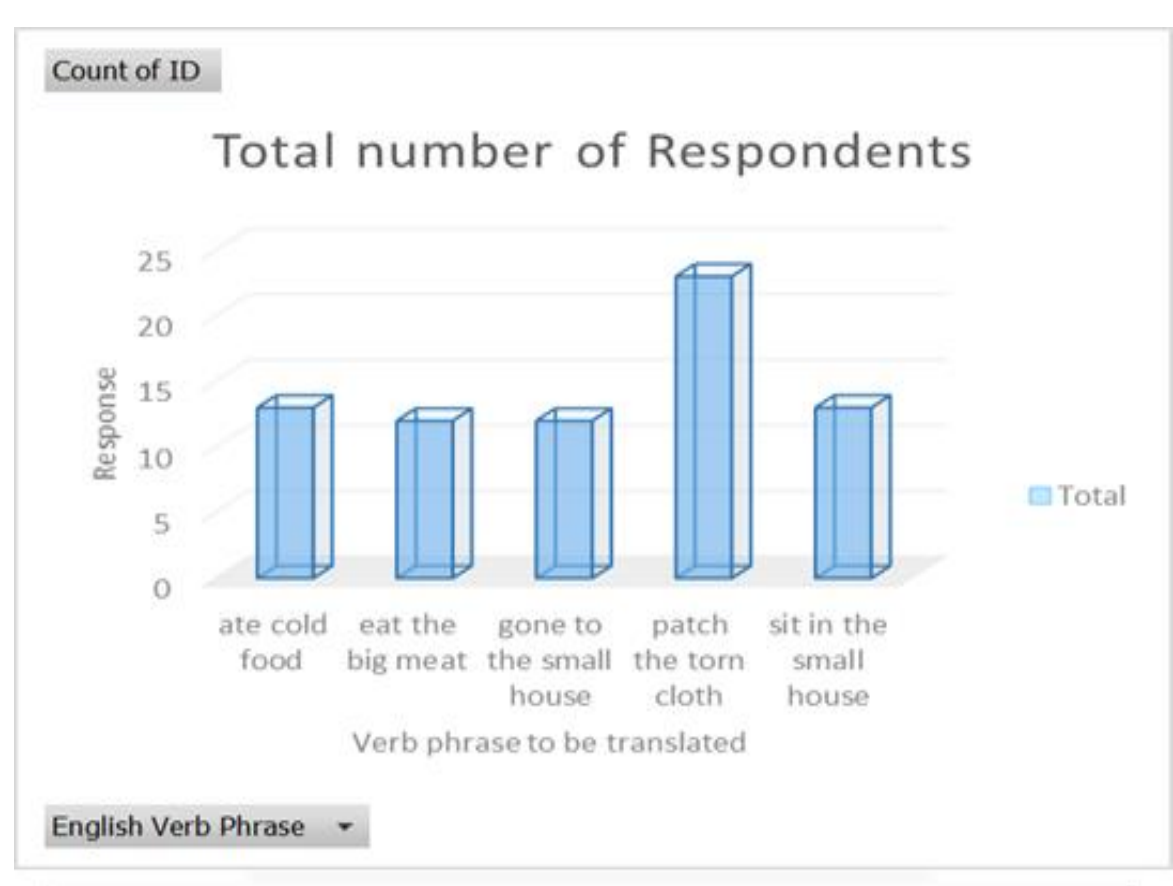

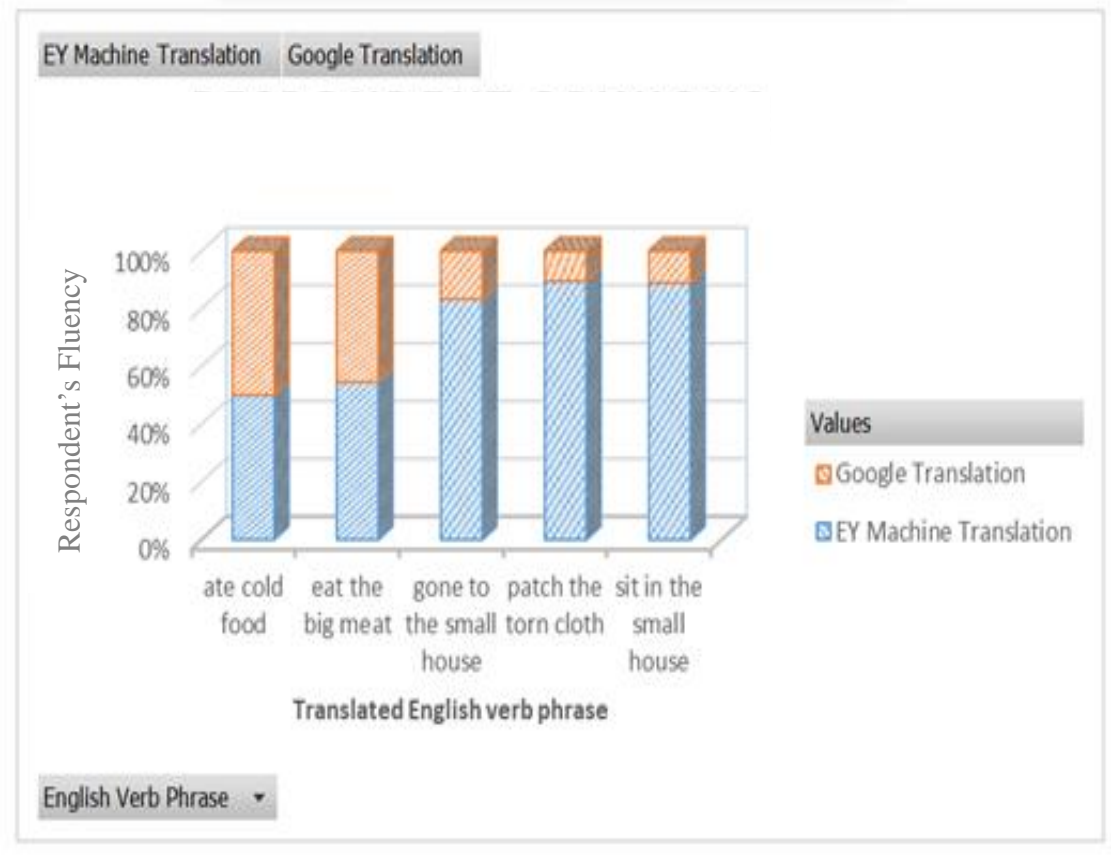

Figure 10: Chart Showing the

## 5 Recommendation and Conclusion

The translation conveys the meaning and undertone of the translation when applicable. The system translates with appropriate tone, marks, and under-dots. This machine translation can be used as a means of teaching students (it is not a close model and practically easy to learn); therefore, a lot of improvements can be made in the part of the phrases that make up the sentence. There is room to make the system's engine more robust by adding to the rewrite rules and more especially looking at improving the system when adverb is part of the Verb phrase as a modifier. And finally, the system can be further improved by adding to the database, this will improve the response of the system given any relevant query.

## Conflict of Interest

No conflict of interest between the authors.

# A Appendices

A complete EY corpus used can be accessed here[2]

[2] https://github.com/BenAji/VerbPhrase/blob/main/EY-Corpus.docx

**Noun**

***Parts of the body***

| | |
|---|---|
| arm | Apa |
| back | Eyin |
| cheeks | Ereke or eeke |
| chest | Aya |
| ear | Eti |

***Objects***

| | |
|---|---|
| bathroom | Baluwe |
| bed | beedi |
| bedroom | Yara Ibusun |
| ceiling | Orile |

***Food/Meal***

| | |
|---|---|
| bread | Buredi |
| breakfast | Onje aaro |
| butter | Bota |

***Yorùbá Numbers (Cardinal and Ordinal)***

| | |
|---|---|
| One | eyokan |
| first | okan |
| two | meji |

**Adjectives**

***Colors***

| | |
|---|---|
| Black | dudu |
| Grey | awo-resuresu |

*Sizes*

| | |
|---|---|
| big | Nla |
| deep | Jin |
| long | Gun or gigun |
| narrow | Tinrin |

# B Appendices

Examples of EY Machine Translation of Simple verb phrase

| English VP Text input | MT Yorùbá Equivalent |
|---|---|
| killed a boy | Pa omodokunrin kan |
| killed the big man | Pa omokunrin nla |
| eat cold food | Je ounje tutu |
| ate the hot food | Je ounje gbigbona naa |
| cook the big meat | Se eeran nla naa |
| go to the small house | Lo si ile kekere naa |
| patch the torn cloth | Gan aso yiya kan |
| sit in the small house | Joko niun ile kekere naa |
| Gave mother the cold water | Fun iya ni omi tutu naa |